\documentclass[default]{sn-jnl}

\normalbaroutside
\usepackage{todonotes}
\usepackage{natbib}
\usepackage{nth}

\jyear{2023}%

\theoremstyle{thmstyleone}%
\usepackage{enumitem}
\setlist[itemize]{labelsep=3mm}

\theoremstyle{thmstyletwo}%

\theoremstyle{thmstylethree}%
\usepackage{subcaption}
\usepackage{array}
\usepackage{tabularray}
\usepackage{makecell}

\begin{document}

\title[EML and Games]{Evolutionary Machine Learning and Games}


\author[1,2]{\fnm{Julian} \sur{Togelius}}

\author[3]{\fnm{Ahmed} \sur{Khalifa}}

\author[1]{\fnm{Sam} \sur{Earle}}

\author[1]{\fnm{Michael Cerny} \sur{Green}}

\author[4]{\fnm{Lisa} \sur{Soros}}

 \affil*[1]{\orgdiv{Computer Science and Engineering}, \orgname{New York University}, \orgaddress{\street{370 Jay Street}, \city{Brooklyn}, \postcode{07960}, \state{New York}, \country{United States of America}}}

 \affil[2]{ \orgname{modl.ai}, \orgaddress{\street{Nørrebrogade 184}, \city{Copenhagen}, \postcode{2200}, \country{Denmark}}}

 \affil[3]{\orgdiv{Institute of Digital Games}, \orgname{University of Malta}, \orgaddress{\street{20 Triq L-Esperanto}, \city{Msida}, \postcode{MSD2080}, \country{Malta}}}

 \affil[4]{\orgdiv{Computer Science}, \orgname{Barnard College}, \orgaddress{\street{3009 Broadway}, \city{New York}, 
  \state{New York},
 \postcode{10027}, \country{United States of America}}}



\abstract{Evolutionary machine learning (EML) has been applied to games in multiple ways, and for multiple different purposes. Importantly, AI research in games is not only about playing games; it is also about generating game content, modeling players, and many other applications. Many of these applications pose interesting problems for EML. We will structure this chapter on EML for games based on whether evolution is used to augment machine learning (ML) or ML is used to augment evolution. For completeness, we also briefly discuss the usage of ML and evolution separately in games.}

\keywords{Games, Procedural Content Generation, Automated Game Playing, Player Modeling, NeuroEvolution}



\maketitle







\section{Introduction}\label{sec1}

Games of all sorts (including card games, board games, and video games) provide a rich domain for exploring computational intelligence. In many ways, games reflect the parts of the real world that we as humans find interesting, isolating key facets of our experience and encapsulating them within a tractable and interactive medium. 

Beyond mere entertainment, games provide a unique domain for exploring and evaluating AI. Historically, the intersection of AI and games has focused on agents that play specific games \emph{well}. In this way, games complement the litany of task environments for AI such as embodied agent control. However, games offer additional challenges beyond reward maximization such as effecting spirited play or emulating the style of particular humans. 

In addition to playing games, there are challenges including generating content (such as levels, quests, textures, and characters), modeling players, matching players, and adapting interfaces. Content generation in particular requires deeply creative algorithms capable of understanding the essence of a domain and conjuring up new artifacts. In this way, games also provide an opportunity for exploring concepts such as algorithmic innovation and open-endedness. 


In this chapter, we survey the application of EML to games. We take an inclusive view of EML, focusing on cases where a ML model is evolved, but including examples of all kinds of interaction between evolution and ML. We also give brief overviews of the use of non-evolutionary ML and non-ML evolution in games, but given the breadth of the topic, those sections are mere sketches.

EML can, in one form or another, be applied to almost any AI challenge in games. However, this family of methods has seen much more application in some areas rather than others. Reflecting on this, a relatively large number of examples in this chapter will be from game content generation. But there will also be plenty of examples of game-playing evolutionary ML.



\section{Machine Learning in Games}

Some of the earliest advancements in ML are due to research on game-playing. In particular, Samuel's Checkers player from 1959~\cite{samuel1959some} was the first example of what we now call reinforcement learning. While programs for Chess~\cite{campbell2002deep}, Checkers~\cite{schaeffer2007checkers}, and Go initially built (and usually still build) on some form of tree search, machine-learned board value functions were introduced at an early stage and became crucial to any advanced efforts to play classical board games. These value functions could be learned through supervised learning, reinforcement learning, or some combination. AlphaGo~\cite{silver2016mastering} and AlphaZero~\cite{silver2017mastering} represent very successful combinations of tree search with ML that originated in research on Go playing but have found applications in a wide variety of fields.

At the outset of the deep learning era, certain video games came to play important roles as testbeds and benchmarks of deep reinforcement learning algorithms. Deep Q-networks were introduced in a landmark Nature paper in 2015~\cite{mnih2015human}, where they were trained on games from the Atari 2600 console via the Arcade Learning Environment~\cite{bellemare2013arcade}. Since then, Atari games have been a crucial benchmark for deep reinforcement learning, and new algorithms are often tested on Atari games first~\cite{justesen2019deep}. Several other video games have also become important as benchmarks for RL algorithms, notably Doom~\cite{kempka2016vizdoom} and MineCraft~\cite{guss2019minerl}. To a lesser extent, supervised learning has been applied to these games to learn game-playing from large sets of human playtraces~\cite{baker2023video}.

Although a much smaller research area, ML algorithms of various kinds have also been applied to generating game content~\cite{summerville2018procedural}. Most of these applications fall into self-supervised learning, such as using generative adversarial networks or recurrent neural networks for generating levels for Super Mario Bros~\cite{volz2018evolving,awiszus2020toad,summerville2016super} or cards for Magic: The Gathering~\cite{summerville2016mystical}. However, there have also been attempts to use reinforcement learning for level generation more recently~\cite{khalifa2020pcgrl}.

Other AI challenges in games for which ML has been used prominently include player modeling~\cite{smith2011inclusive}, cheat detection~\cite{hong2006identification}, and matchmaking~\cite{minka2020machine}. In player modeling, preference learning has been successfully applied to predicting player affect in response to game levels, game situations, or behavior. For cheat detection, explainable ML algorithms (such as decision trees) are most commonly used as the AI cannot act by itself and block players. It usually needs the moderator's input to confirm if the flagged player is a cheater before banning the player from the game.  Microsoft's TrueMatch system, widely used in XBox games, uses ML on top of a ranking system to learn to match players well.


\section{Evolution in Games}

The robustness and wide applicability of evolutionary search means that evolutionary algorithms have been applied rather extensively to games.
Many, perhaps most, examples of applications of evolution to games are in the form of neuroevolution, which counts as EML. However, there are also ``pure'' applications of evolutionary computation in games. Here we will survey two types: evolutionary planning and search-based procedural content generation.

\subsection{Evolutionary planning}

For playing a game that has a fast-forward model, one preferably uses some kind of search algorithm for planning. Traditionally, this would mean a tree search algorithm, such as a Minimax variation for two-player games or some form of A* for single-player games~\cite{russell2010artificial} For games with a nontrivial search depth, one would need to cut off the search at some specified depth and use a board or state evaluation function to evaluate the deepest search node. 

The plan that is generated by the tree search algorithm is simply a list of actions, with a numeric value (obtained from the state evaluation function) that signifies the projected value of that sequence of actions. Recognizing this, one might just as well use an evolutionary algorithm to evolve the plan, using the state evaluation function as a fitness function. In an approach called rolling horizon evolution~\cite{perez2013rolling}, several groups have done this for single-player games~\cite{gaina2017rolling}, with results that are generally competitive with state-of-the-art tree search algorithms such as Monte Carlo Tree Search~\cite{browne2012survey}

The advantages of evolution over classic tree search algorithms become more clear in games with a very large branching factor~\cite{sfikas2021playing}. Games that feature control over multiple units, including many real-time or turn-based strategy games, can have branching factors in the millions or even billions. This chokes most tree search algorithms as the search never gets past the first few moves. Evolutionary algorithms, operating on the level of the whole plan as a sequence, are not affected by this limitation. Results for several strategy games~\cite{justesen2016online} show that evolutionary planning can out-compete tree search by a large margin.

\subsection{Search-based PCG}

After game playing, procedural content generation (PCG) is probably the topic in games that has received the most interest from the AI research community~\cite{shaker2016procedural}. Within academic research on PCG, evolutionary computation is currently the dominant approach because of the natural fit between the method and the problem. In search-based PCG, as the application of evolution to PCG problem is called, individual pieces of game content are evolved with a fitness function that reflects some measure of content quality~\cite{togelius2011search}. This approach has been applied to a large number of types of game content, including quests~\cite{de2019procedural}, music~\cite{eigenfeldt2012corpus}, textures~\cite{wiens2002gentropy}, rules~\cite{cook2013mechanic}, and levels for platform games~\cite{shaker2012evolving}, first-person shooters~\cite{cardamone2011evolving}, and real-time strategy games~\cite{liapis2013generating}.

Finding a fitness function that accurately reflects the quality of a candidate content artifact is typically nontrivial. Judging the quality of, e.g., a game level is inherently hard and often requires playing the game. Writing code that performs this quality judgment is often harder still. Therefore, many search-based PCG implementations use simulation-based evaluation, where an agent of some kind plays part of the game that incorporates the content under evaluation. For example, the agent could play a level to see if it is playable under certain conditions~\cite{khalifa2019intentional}. Such agents may be based on ML~\cite{togelius2008experiment}. In other cases, the fitness function is model-based and relies on some kind of learned model to do the evaluation~\cite{karavolos2018using}. In yet other works, fitness is assigned interactively by a human player~\cite{gallotta2023ple}.

\section{Evolutionary Machine Learning in Games}

\begin{figure}
    \centering
    \includegraphics[width=\linewidth]{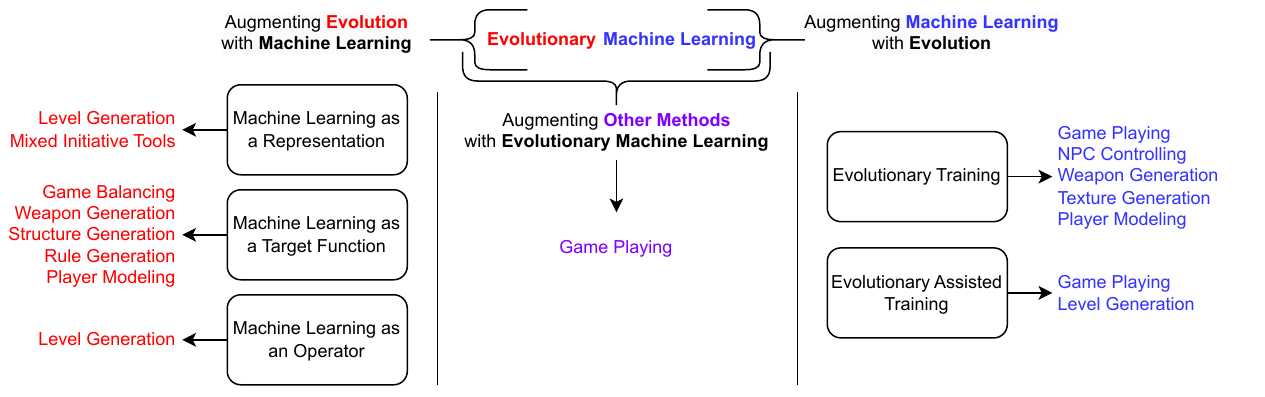}
    \caption{Taxonomy of EML in Games based on the dynamics between evolution and ML with examples of the usage of these techniques in games.}
    \label{fig:taxonomy}
\end{figure}

In this section, we overview the use of EML in games research. While one way of categorizing this body of work is to divide games application (e.g. planning for non-player character actions, PCG, etc.), another is to tear apart EML, considering the ways in which evolution and ML interact. We take the latter approach in the context of this book, as the former is better suited to works on AI and Games~\cite{yannakakis2018artificial}.
We survey research in games using EML (as shown in figure~\ref{fig:taxonomy}) by:
\begin{itemize}
    \item Augmenting Machine Learning with Evolution.
    \item Augmenting Evolution with Machine Learning.
    \item Augmenting other methods with Evolutionary Machine Learning.
\end{itemize}
The way we acquired this division is by looking into what is the end goal of the work that uses EML and analyze what is the core algorithm that is being used. For example, if the end goal is to have a neural network agent that controls a player in a video game, then evolution in that case is helping the ML (neural network) to achieve its goal. On the other hand, if the goal is to use evolution to find a certain content, then the search process is what is important and the ML is helping it to achieve its goal. We also found some other cases where the goal algorithm is neither an evolutionary algorithm nor ML, but where EML nonetheless plays an important supporting role.

\section{Augmenting Machine Learning with Evolution}
The easiest way to incorporate the evolutionary algorithm with ML is to utilize evolutionary optimization power to help the ML algorithm. Thinking about it from this direction allows us to see two ways where evolution can help. The first thing is to use the evolutionary algorithm as the training process for the ML or the evolutionary algorithm is adjusting and helping the normal training process of ML.

\subsection{Evolutionary Training}\label{sec:evo_train}

Evolution can be used to directly train neural networks, either the weights of the network, its topology, or both. This is called \emph{neuroevolution} and has a long history in artificial life, robotics~\cite{nolfi2000evolutionary}, and video game research~\cite{risi2015neuroevolution}. It should be noted that some recent neuroevolution work, specifically focused on finding structures for deep networks, is alternatively billed as ``neural architecture search''~\cite{elsken2019neural}.

\subsubsection{Neuroevolution for Gameplay Tasks}

The probably most common use for neuroevolution in games is as a form of reinforcement learning. Here, the fitness evaluation of a chromosome is calculated by converting the genotype to a neural network (which might be as simple as setting the weights of a fixed topology) and using it to try to play the game, and basing the fitness on how well the network played the game. Compared to other forms of reinforcement learning, such as those based on variations of temporal difference learning or policy gradients, neuroevolutionary reinforcement learning tends to be more reliable but has issues with scaling to very large networks~\cite{togelius2009ontogenetic,risi2015neuroevolution}.

In board games, the most common (and probably most sensible) version of neuroevolution is to evolve a board value function used together with some version of the Minimax algorithm. Successful applications of this method go back a few decades~\cite{fogel2002blondie24}; see also section~\ref{sec:augmenting}.

In video games, where fast and deterministic forward models are rare, it is typically more appropriate to use neuroevolution to generate a network that directly selects what action to take based on an observation of the environment. Early examples of this type of neuroevolution include work on 2D car racing games~\cite{togelius2005evolving,togelius2006evolving}. In those examples, the neural networks were fed simulated rangefinders representing the track geometry in front of the car, and outputted accelerator/brake and steering commands. Another relatively early example of this type of neuroevolution is the evolution of neural nets to play Super Mario Bros based on discrete grid sensor inputs~\cite{togelius2009super}.

As the ALE benchmark suite, featuring games from the old Atari 2600 video game console, became popular with the rise of deep reinforcement learning, evolutionary algorithms were also applied to learn to play Atari games from pixel inputs. An influential paper by a team at OpenAI showed that a simple evolution strategy can be competitive with standard DQN, and in particular that evolution scales effortlessly compared to other types of reinforcement learning~\cite{salimans2017evolution}. Follow-up work by other authors showed that the slightly more sophisticated Canonical Evolution Strategy can perform much better~\cite{chrabaszczback18back}. However, the more recent and sophisticated gradient-based RL methods outperform existing evolutionary methods on Atari games. In general, while standard neuroevolution remains competitive with gradient-based RL when the network size is small, it tends to not scale very well to very large networks. This may be because the single fitness measure imparts less information than the dense rewards that can be used with some RL methods, or because of the lack of directional information in the random mutations in a high-dimensional space. 

Various attempts have been made to overcome this limitation of neuroevolutionary reinforcement learning for games with high-dimensional (such as visual) input space. One idea is to separate visual processing from the policy. This way, a smaller network that receives a lower-dimensional encoding of the observation can be trained effectively by evolution. One idea is to use an autoencoder to compress the visual input, feeding the smaller policy network of the bottleneck layer in the autoencoder. This was tested successfully on the classic FPS game Doom, in a setup where the autoencoder was continually re-trained as new sections of the level were discovered~\cite{alvernaz2017autoencoder}. Another idea is to represent the state using a dictionary of centroids in state space, feeding a sparse encoding to the policy network. This setup can be used to evolve surprisingly tiny networks, with as few as six neurons, to play Atari games well~\cite{cuccu2018playing}.

One of the areas where neuroevolution can do things that regular reinforcement arguably cannot is in topology and weight evolution, where both the topology and the weights of the network are evolved. The pre-eminent algorithm here is NEAT by Stanley and Miikkulainen, which has been applied to learn to play various games and performs very well on small to moderate-size networks~\cite{stanley2002evolving}. Later, it has been applied to the General Video Game framework with decent results~\cite{perez2020rolling}. HyperNEAT is an indirect encoding that can in principle handle much larger input spaces~\cite{stanley2009hypercube}, and which has been applied to playing Atari games with some success~\cite{carvelli2020evolving}.

\subsubsection{Neuroevolution for Non-Gameplay Tasks}

Neuroevolution can also be applied to other tasks in games besides game playing. In particular, there are several prominent examples of neuroevolution for procedural content generation. This includes the Galactic Arms Race game~\cite{hastings2009automatic}, where players collaboratively evolve weapons, and the Petalz social network game where players evolve flowers~\cite{risi2012combining}. One can even use neuroevolution as a form of meta-content generator, evolving a population of networks that can generate a large range of different levels~\cite{earle2022illuminating}. An interesting use of neuroevolution for non-player characters is in the NERO game, where players train populations of agents to fight for them~\cite{stanley2005real}. Even earlier, the cult video game Creatures uses a form of neuroevolution as a game mechanic.

Neuroevolution has also been used to to model a certain playstyle~\cite{holmgaard2014evolving,holmgaard2016evolving}. Holmgaard et al. used a $\mu + \lambda$ evolutionary strategy to evolve a single-layer neural network that decides on the next target in MiniDungeons. The agent then uses the A* algorithm to navigate toward the selected target. The agents were evolved using action agreement ratio~\cite{holmgaard2014evolving}, tactical agreement ratio, and strategic agreement ratio~\cite{holmgaard2016evolving} to mimic different playstyles from a corpus of collected human data. The final evolved agents were not only able to finish unseen levels but also navigate them in the same vein as the target playstyle.
 





\subsection{Evolutionary Assisted Training}
In the work described in section~\ref{sec:evo_train} above, the evolutionary process serves to learn more sophisticated policies -- e.g., for game playing or design, whether by the modification of action sequences, game assets, or of neural network weights or architectures. In this section, on the other hand, the learning or training process is implemented by some non-evolutionary algorithm (typically gradient descent), while evolution exists as an auxiliary process that renders learning more capable or efficient. This section focuses primarily on a particularly prominent role taken on by such an assistive evolutionary process, which is creating or curating a ``curriculum'' for a learning agent.


\subsubsection{Curriculum Design in Machine Learning}

 Curriculum learning~\cite{bengio2009curriculum} takes inspiration from human developmental psychology, where, by way of example, children are taught complex subjects (e.g. mathematics) incrementally over time, by first familiarizing them with more elementary and fundamental concepts (such as basic arithmetic) before moving on to more sophisticated techniques (such as calculus or linear algebra).
In the context of ML, a curriculum may expose a learning agent to increasingly complex training examples, switching from simpler to more complex ones once the agent's abilities reach a certain level.

For example, a neural network-based agent tasked with navigating through a maze to a goal tile will in all likelihood be incapable of navigating a complex maze by sheer luck at the very beginning of training (i.e. with randomly initialized weights resulting in it taking effectively random actions at each step). It would thus be wasteful to expose it to such complex mazes at the beginning of training. It has a much better chance, on the other hand, of stumbling upon the goal tile in simpler mazes, where the goal is close to the starting position. A curriculum learning approach could thus order mazes by their complexity (or the distance between start and goal positions) only introducing more complex mazes into the pool of training data once the agent has learned to reliably solve a set of simpler mazes. 


In ML, curriculum learning can not only make training more efficient (avoiding training on infeasible examples early on) but also lead to improved generalization, by allowing the model to incrementally learn more meaningful representations of the problem at hand.
The importance of curriculum learning for training generally capable strategies over a large space of tasks has been demonstrated in XLand \cite{team2021open,team2023human}, where a large set of tasks are generated in a 3D environment, where embodied agents take visual input and must navigate procedurally-generated terrain to achieve certain goal states, which involve manipulating and re-combining various primitive objects. In XLand, a massive space of tasks (involving different rules and initial map conditions) are generated before learning, then curated during agent training so as to produce a curriculum of increasing complexity. This curation is aided by metrics using agent play-through to measure whether a given task is both non-trivial and learnable (i.e. whether it is on the ``frontier'' of agent abilities). The resulting player agents are able to generalize to new tasks -- involving different object-recombination mechanics and goal states -- that were not seen during training.

\subsubsection{Evolutionary Curriculum Design}

Other work generates training environments on the fly, then curates or filters them to ensure learnability, resulting in a series of environments of increasing complexity over the course of agent training player agents. PAIRED~\cite{dennis2020emergent} introduces this notion of ``Unsupervised Environment Design'', using an RL agent to generate level layouts of training environments so as to maximize learnability (formulated as approximate regret). More recently, ACCEL~\cite{parker2022evolving} effectively evolves these environments, applying small mutations to level layouts and filtering them for regret. The domains considered include tile-based maze-like environments in which the player must navigate to a goal (while sometimes avoiding lava) and the 2D physics-based ``bipedal walker'' environment. The mutations involve changing the state of given tiles in the former case and changing the height of the terrain in the latter. The fitness criterion of a mutated level is the regret it induces in the player agent. The authors show that the evolution of increasingly complex (lengthy and obstacle-heavy mazes, and rough or steep terrain) coincide with the development of a robust player agent capable of generalizing to unseen environments (e.g. capable of solving a maze in a grid much larger than those evolved during training).

POET~\cite{wang2019paired} similarly evolves environments online while training player policies in the bipedal walker environment, though it trains multiple player agents concurrently, pairing these with particular environments (and occasionally transferring players to different environments mid-training in a paradigm reminiscent of transfer learning \cite{torrey2010transfer}). Here, terrain is represented by a Compositional Pattern Producing Network (CPPN~\cite{stanley2007compositional}), and evolved using Neuroevolution of Augmenting Topologies (NEAT~\cite{stanley2002evolving}). Agent policies are optimized via an Evolutionary Strategy~\cite{salimans2017evolution}. Mutated environments are added to the pool of environments if they are neither too hard nor too easy for existing agents, and if they are sufficiently novel with respect to environments already in the pool, thus encouraging diversity among the problems generated (and solved) by the algorithm.
PINSKY~\cite{dharna2020co} applies the framework introduced by POET to 2D tile-based games.

In Go-Explore~\cite{ecoffet2021first}, a variant of novelty search~\cite{lehman2011novelty} is used to incentivize exploration by an RL agent, by effectively maintaining an archive of behaviors that result in novel states in a sparse reward problem. The domain tackled in this work is Montezuma's Revenge, a side-scrolling adventure game from the Atari 2600 suite, which was formerly unsolved by methods not relying on expert demonstrations. Go-Explore facilitates exploration by looking at the $(x, y)$ coordinate location of the player avatar on the screen, and, whenever a new position is reached by the player, storing in memory the sequence of behaviors that led to this novel state. On subsequent play-throughs, the agent can then reliably navigate back to an $(x, y)$ position on the frontier of that which it has already explored, so that it will tend to explore further and further out from its starting location, and so that once a tricky segment has been solved -- such as jumping carefully across a dangerous chasm -- it can be reliably solved on every subsequent play-through without further trial and error. When the goal state is reached, they train a game-playing agent on the successful trajectories from the archive.


\section{Augmenting Evolution with Machine Learning}

The goal in augmenting the evolutionary process is to mimic the output of the evolution process by using ML to create an easier landscape to navigate, better operators, faster fitness function, etc.

\subsection{Machine Learning as a Representation}

\begin{figure}
    \centering
    \begin{subfigure}[t]{0.25\linewidth}
        \centering
        \includegraphics[width=\linewidth]{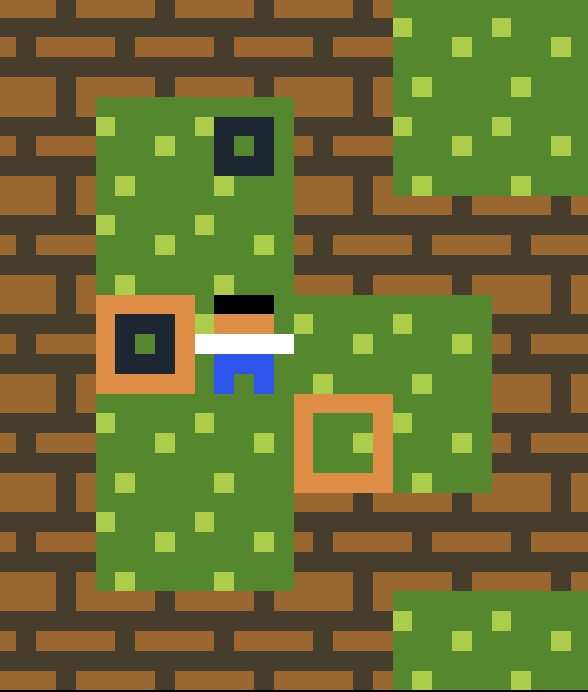}
        \caption{}
        \label{fig:sokoban_with}
    \end{subfigure}
    \begin{subfigure}[t]{0.25\linewidth}
        \centering
        \includegraphics[width=\linewidth]{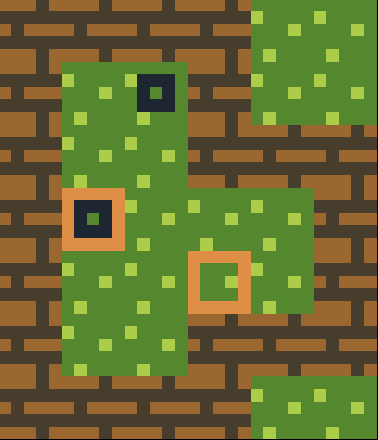}
        \caption{}
        \label{fig:sokoban_without}
    \end{subfigure}
    \caption{An example of a Sokoban level with and without the human player.}
    \label{fig:sokoban_player_example}
\end{figure}

Evolutionary algorithms are sensitive to the input representation~\cite{golberg1989genetic}. There is a body of research on understanding and exploring the different representations and their effects~\cite{rothlauf2006representations}. In most of these works, it is agreed that a good representation provides us with a smooth fitness landscape. This means that individuals that are close to each other in representation space have similar fitness. In many game domains such as level generation, a direct representation (2D matrix of tiles) is usually not the best as it is too sensitive to small changes. For example, removing the player tile from a level will make the level unplayable immediately (figure~\ref{fig:sokoban_player_example}). Instead of utilizing domain knowledge to figure out a good representation, ML models can be used to learn this representation. In that case, ML acts as a genotype-to-phenotype mapper for the evolutionary algorithm. The evolutionary algorithm can just work on the provided representation (called latent variables) and modify it using a mutation function then the ML transforms it to the phenotype where it gets evaluated using a fitness function.

\begin{figure}[]
  \resizebox{\linewidth}{!}{
  \begin{tabular}{l|cc}
    \multirow{2}{*}{\Large Level 1-3} & \includegraphics[width=\linewidth]{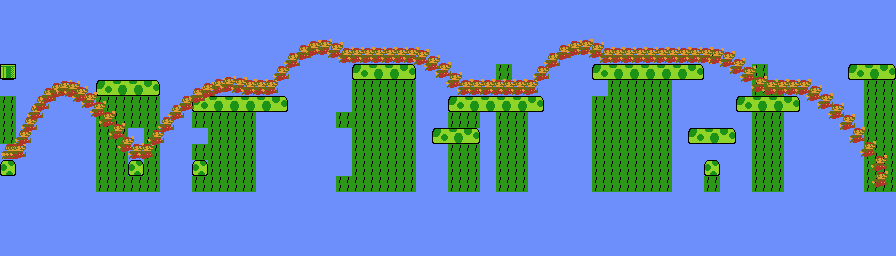} & \includegraphics[width=\linewidth]{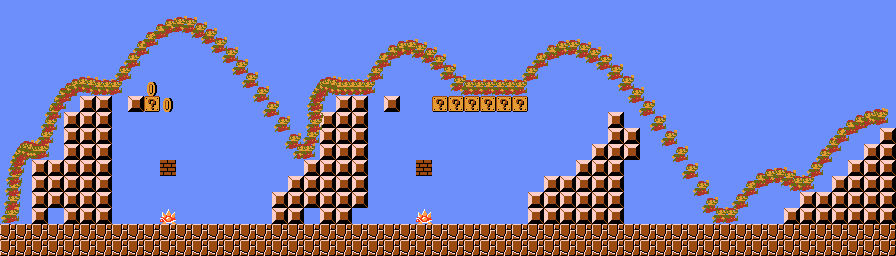} \\
    & \includegraphics[width=\linewidth]{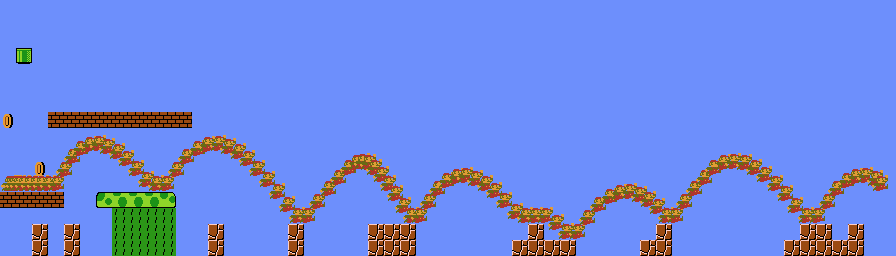} & \includegraphics[width=\linewidth]{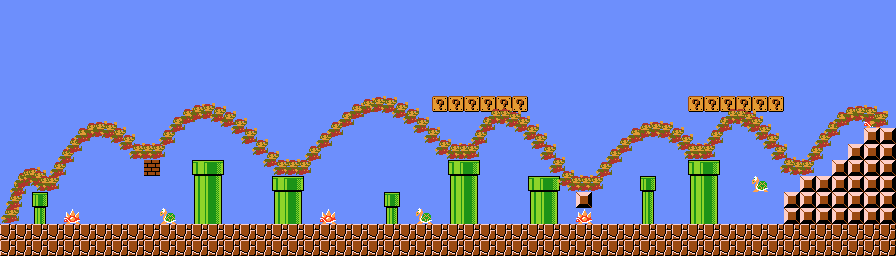} \\\hline
    \\
     & \multicolumn{2}{c}{\Large Level 1-1}\\
  \end{tabular}
  }
  \caption{Generated Super Mario scenes using CMA-ME that are similar to either sky levels or ground levels.}
  \label{fig:KL-levels}
\end{figure}

\subsubsection{Super Mario Bros.}

Although there are many ML methods that can be used to learn a good representation, most of the work focuses on either AutoEncoders~\cite{kramer1991nonlinear} or Generative Adversarial Networks (GANs)~\cite{goodfellow2020generative}. For example, Volz et al.~\cite{volz2018evolving} trained a GAN on the levels of Super Mario Bros. (Nintendo, 1985). The levels are divided into single scenes of size (28x14) using a sliding window. After training the network on generating new scenes, they used CMA-ES~\cite{hansen2003reducing} algorithm to search the latent space of the GAN for playable Mario levels with different game features such as maximizing/minimizing the number of jumps. The fitness was calculated using an A* agent that measures the playability percentage of the generated levels and the number of jumps that have been performed. The generated levels mostly follow the structure of original Mario levels without the need for specifying that in the fitness function or the representation~\cite{shaker2012evolving,dahlskog2013patterns}. Following the previous work, Fontaine et al.~\cite{fontaine2021illuminating} used a quality diversity algorithm called Covariance Matrix Adaptation MAP-Elites (CMA-ME)~\cite{fontaine2020covariance} to discover new and diverse playable levels. Utilizing CMA-ME helped the authors overcome the repetition of the generated levels using normal CMA-ES and find new levels that are hard to find using normal CMA-ES such as levels that combine two different level styles (figure~\ref{fig:KL-levels}). Another advantage is at the end of the run, the algorithm manages to produce a corpus of playable levels with different features instead of a single playable level as in the case of CMA-ES.

\begin{figure}
    \centering
    \includegraphics[width=\linewidth]{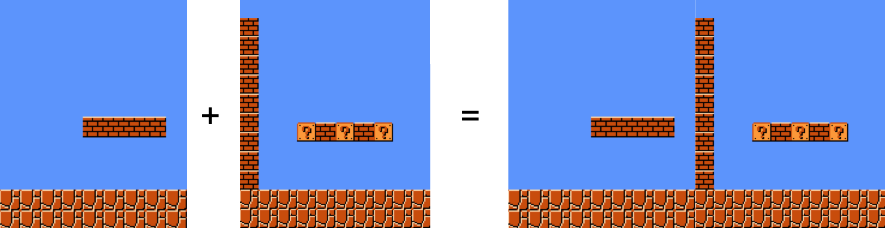}
    \caption{These Mario scenes are playable on their own, but combining them in that order will not be playable.}
    \label{fig:mario_scenes}
\end{figure}

So far the previously discussed work only focused on generating a small scene of Super Mario levels of size (28x14), creating a full level by concatenating the generated scenes sometimes might end in an unplayable level (see figure~\ref{fig:mario_scenes}). This is because evolution hasn't any context about the previously generated scenes. This can be easily solved by modifying the fitness to incorporate all the previous scenes. The problem is that evolution cannot change anything from the previously generated scenes. This might cause evolution to repeat certain patterns to solve the problem. Another simple solution is to evolve the whole level all at once and use the GAN to transform each sequence independently. The problem with that is for very long levels, this could be hard as the representation might be big and not easy to evolve due to conflict in fitness between different parts of the chromosome. To solve this issue, Schrum et al.~\cite{schrum2022hybrid} proposed encoding the big level using a compressed representation which is encoded in the form of a CPPN~\cite{stanley2007compositional}. The CPPN takes the location of the scene and it spits out the latent variable for that area. The NEAT~\cite{stanley2002evolving} algorithm was used to create the full level. The output of this experiment shows that using CPPNs to evolve the structure of the level followed by normal evolution on the concatenated latent variables produced better results than any part alone.

\subsubsection{Other Applications}

Tanabe et al.~\cite{tanabe2021level} used latent variable evolution with Variational AutoEncoder (VAE)~\cite{rezende2014stochastic} to generate levels for Angry Birds (Rovio, 2009). They treated the generated levels as a sequence instead of being a tile map to overcome the problem of having variable size objects and having objects overlayed each other. They searched the latent space of VAE using CMA-ES algorithm to find levels that minimize the usage of birds, maximize the number of pigs, etc. 
Another avenue is to use interactive evolution to create a mixed initiative tool~\cite{shaker2016mixed} combined with the power of ML. This is usually a common avenue in subjective domains that are hard to measure its success. There is a substantial breadth of work in this area, and a full review is outside the scope of this book chapter. However, some recent works generate art~\cite{simon2018artbreeder}, human faces~\cite{zaltron2020cg}, backgrounds, and shoes~\cite{bontrager2018deep}. However, as far as we know, these tools have not been used in texture generation or game art. This could be an interesting avenue to explore with the rise of large generative models such as Stable Diffusion~\cite{rombach2021highresolution}.


\subsection{Machine learning as a Target Function}
Usually, the bottleneck for evolution is calculating the fitness function~\cite{sobester2008engineering}, which motivated people to seek methods to approximate its value to speed up the evolutionary process. This approach involves surrogate models~\cite{sobester2008engineering}. Surrogate models have been modeled using different ML methods such as linear regression~\cite{gravina2016surprise}, gaussian mixture models~\cite{liu2013gaussian}, neural networks~\cite{karavolos2018using}, etc. In this section, we decided to call it a target function instead of a fitness function as ML could be used to approximate not only the fitness function but also features of the phenotype such as time to finish the level, number of jumps in a playtrace, etc. These features can be used either as behavior characteristics in quality diversity algorithms~\cite{zhang2022deep,bhatt2022deep} or a diversity score for divergent search algorithms~\cite{liapis2021transforming,barthet2022open}.

We can divide surrogate models based on how they are trained: online and offline methods. The online method focuses on training the surrogate model as a part of the evolutionary process. The evolution helps to create the dataset for training the ML model. On the other hand, the offline method focuses on training the ML model beforehand. This makes surrogate models similar to reinforcement learning where the online method is similar to off-policy reinforcement learning and the offline method is similar to offline reinforcement learning where we collect the data before training the model.

\subsubsection{Online Methods}

In the games domain, most of the work utilizes online training due to the small number of data that can be found to train ML models. Volz et al.~\cite{volz2016demonstrating} used multi-objective optimization to create balanced cards for the Top Trumps card game (Dubreq, 1978). The goal is to generate a group of unique balanced decks such that there is no dominant strategy to win the game. The surrogate model here is a statistical model that predicts the minimum number of simulation that is needed to have an accurate estimation of the win rate. The results showed that generated decks from surrogate models are as good as normal simulations and require a lot less computational power. In a similar vein, Zhang et al.~\cite{zhang2022deep} used Deep Surrogate Assisted MAP-Elites to generate a deck of cards for Hearthstone (Blizzard, 2014). They not only used the surrogate model to calculate the fitness (the average difference of health between both players) but also the behavior characteristics (the average number of cards in hand and the number of turns till the end of the match) needed for the MAP-Elites Archive. The model is trained online after a fixed number of iterations to make sure the output is correct and up to date. Bhatt et al.~\cite{bhatt2022deep} generalized the system by adding another surrogate model that predicts the agent playtrace instead of just the final metrics and tested it on generating mazes and Super Mario Bros levels. The new system works better in comparison with the previous one introduced by Zhang et al.~\cite{zhang2022deep} as the introduced agent prediction helps improve the results.

Some quality diversity and divergent search algorithms need ML as their core element such as Surprise Search~\cite{gravina2016surprise}. Surprise search abandons objectivity for the sake of surprise. The surprise score needs to be modeled using a ML algorithm so the evolution can predict new elements. The authors modeled the surprise score using linear regression to predict the genotype of the next generation given the previous generation. In this case, online models are being used as the model gets updated after each generation. Later, Gravina et al.~\cite{gravina2016constrained} adjusted the surprise search algorithm to maintain quality and not only focus on surprise. They utilized the new algorithm (Constrained Surprise Search) to generate a diverse set of balanced weapons for FPS shooter games. They also showcased that it can be used to generate robot controllers,  maze solutions, and new mazes~\cite{gravina2018quality}. Liapis et al.~\cite{liapis2021transforming} and Barthet et al.~\cite{barthet2022open} used a denoising autoencoder to help Constrained Novelty search~\cite{lehman2011abandoning} to find innovative spaceship designs and MineCraft (Mojang Studios, 2009) buildings respectively. The denoising autoencoder is trained online from the found data and then the compressed representation is used to measure the novelty of the generated content.

\subsubsection{Offline Methods}

On the offline side, Karavalos et al.~\cite{karavolos2018using} trained a neural network to predict the win rate and time to finish an FPS match of given a level and the player classes. They trained the model offline on tons of random matches using random levels and random player classes then later used the trained model as a surrogate model to evolve different player classes. Migkotzidis and Liapis~\cite{migkotzidis2021susketch} used the same model as part of a mixed-initiative tool~\cite{yannakakis2014mixed} that can be used to design balanced levels for FPS games. On the side of affective computing, Barthet et al.~\cite{barthet2022generative} use a variant of Go-Explore called Go-Blend  (a quality diversity algorithm that keeps track of the best solutions) to explore game trajectories that can mimic different human play styles and arousal levels. They used a simple K-NN algorithm over the AGAIN dataset~\cite{melhart2021affect} to predict the arousal levels during playing a car racing game. Similarly, Shaker et al.~\cite{shaker2010towards} trained an offline surrogate model on human preference over Mario levels and used it to generate new levels for Super Mario Bros (Nintendo, 1985). Since the search space was not huge, an exhaustive search was used instead of evolution. We included this research as the exhaustive search can be easily replaced with evolution in more complex games.

A different way to use the ML model is to use it to play games instead of directly measuring playability or the attributes and then extracting the needed statistics from the playthrough. For example, Togelius and Schmidhuber~\cite{togelius2008experiment} evolved arcade game rules using $\mu + \lambda$ evolution strategy to evolve the game rules such that the evolved games are fun to play by humans. Togelius and Schmidhuber use Koster's theory of fun~\cite{koster2013theory} to estimate how learnable these games are. The learnability is approximated by measuring the improvement of a neural network agent that is trained using reinforcement learning. The generated games from that experiment were interesting but not fun for humans. This is due to the fact that AI agents play differently than humans and we need models that can model the human experience.

Finally, surrogate models can also be used to assist game-playing agents. Olesen et al.~\cite{olesen2021evolutionary} tried to use EML to control the player in a car racing experiment. ML was used to learn a transition model where given a certain state and action what will be the next state. This process is called a World Model~\cite{ha2018world}, where we try to learn an approximate forward model of a specific environment. In their work, Olesen et al. used a VAE followed by Mixture Density Network~\cite{bishop1994mixture} to learn that forward model. Later, they used an evolutionary planning algorithm (Random Mutation Hill Climber (RMHC)) to control the player car to play the game as efficiently. The system first gets trained using collected data from random policy then later it gets improved using frames and output from the RMHC algorithm. Similarly, Dockhorn et al.~\cite{dockhorn2019learning} used hashmaps and decision trees to learn a local forward model for Sokoban. A local forward model is a model that only cares about the local observation around the player character. They use RHEA to play the game using the local forward model. Although the local forward model has high accuracy, the local model propagates errors over time which causes the planning agent to not achieve very high scores in playing the games.

\subsection{Machine Learning as an Operator}

Models trained via ML can be used as mutation or crossover operators during evolution. This is particularly useful when the space of mutations is very large, making it unlikely that random perturbations will lead to meaningful changes in the genome. In such a scenario, a ML-based model can be trained beforehand so as to learn useful priors over the space of possible mutations. This is particularly appealing when a large dataset of human examples is available.

 As an example, ML models can be trained on mutation trajectories~\cite{lehman2022evolution,khalifa2022mutation}, i.e. the modifications made to an entity over the course of evolution. The end result is a model that can generate levels by modifying levels similar to an evolutionary/search-based generator.

In another example, the problem of genetic programming is highly complex. A very naive approach could be to try mutating a piece of code at the string or character level, but this would result in a combinatorially explosive action space for the mutation operator. For this reason, most genetic programming approaches will ``bake in'' some useful priors into the action space afforded to the operator -- for example by providing a higher-level set of available actions that may involve adding functional blocks of code such as if/else statements instead of individual characters or words, while treating the code as a graph or tree structure instead of merely a string.

Large Language Models (LLMs) have shown impressive performance on next-token prediction tasks when trained auto-regressively on massive corpora of human text scraped from the internet. Despite their simple training scheme, they can be observed during inference to generate text that is coherent at a high level, and with the right prompting, can do things like generating rhyming poems about a particular subject matter in a particular style and answering questions while maintaining long-range dependencies. They have also been fine-tuned on corpora of code, and incorporated into tools such as Github Copilot.

In Evolution through Large Models, the authors take advantage of these learned priors, and prompt LLMs to produce embodied agents capable of traversing the terrain in the SodaWorld environment. The LLM used is a diff model, which is trained on a dataset of code changes and their corresponding natural language descriptions in commit messages.
ELM uses this diff model as a mutation operator by randomly selecting from a fixed set of generic commit messages (i.e. ``made changes to the code'').
This operator is then used inside a quality diversity (QD) loop, that searches for a diverse group of ambulating soft robots in SodaWorld.
The LLM can be additionally fine-tuned on the series of mutations accepted to the archive of elites in QD (an approach similarly applied to environment generation in~\cite{khalifa2022mutation}).


\begin{figure}
    \centering
    \includegraphics[width=1\linewidth]{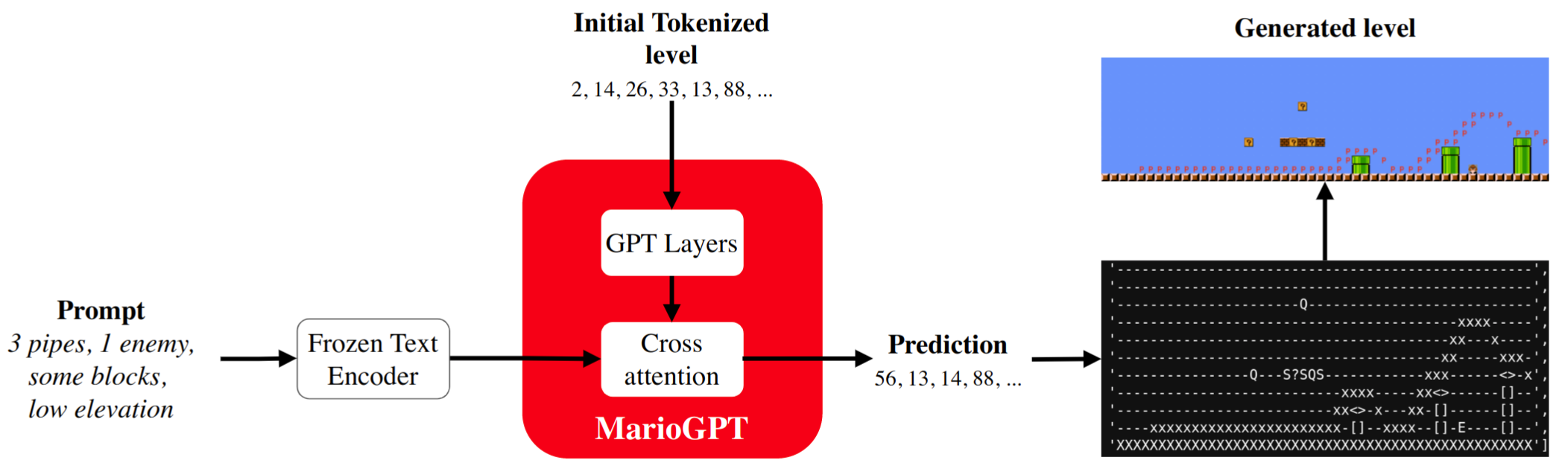}
    \caption{The MarioGPT level-generation pipeline.}
    \label{fig:marioGPT_arch}
\end{figure}

An ELM type approach has also been applied to game levels in MarioGPT~\cite{sudhakaran2023mariogpt}, where GPT-based language models are fine-tuned on a small dataset of Super Mario Bros (Nintendo, 1985) levels. After fine-tuning the language model, novelty search is used as an outer evolutionary loop to search for a diverse set of novel levels generated by the model. Whereas QD fills an archive spanning over multiple behavior characteristics or measures of interest, novelty search simply looks for sufficiently ``different'' individuals, with respect to those already in the archive of elites, by computing the mean distance using some distance function in phenotype space. In MarioGPT, the authors consider the mean distance between the paths corresponding to (tree search-generated) solutions for generated levels, where paths are lines over the 2D plane corresponding to the shape of generated levels.

The previous examples are the only ones that we could find specifically on games. It is clear that this area provides ample opportunities for future exploration. For example, one might consider using LLMs for Neural Architecture Search for game-playing agents similar to the work by Chen et al.~\cite{chen2023evoprompting}. 
Chen et al.~\cite{chen2023evoprompting} represented the neural network architecture as a piece of code that defines layers, skip connections, activations, etc. They used the diff model as a mutation operator to produce new networks. These nets are then evaluated by training them on a labeled dataset (e.g. image classification) and evaluating their test accuracy.


\section{Augmenting Other Methods with Evolutionary Machine Learning}\label{sec:augmenting}
Finally, EML can be used to support other algorithms like tree search and reinforcement learning. Most of the known work focused on using evolution to find/train neural networks that can support tree search algorithms. Blondie24~\cite{fogel2002blondie24} is an AI agent that can play checkers very efficiently. It was able to beat 99.61\% of the matches that it played against 165 human players. The algorithm uses the Min-Max tree search algorithm~\cite{russell2010artificial} with the support of a neural network as a state evaluator. The neural network was trained using an evolutionary algorithm. In a similar manner, Reisinger et al.~\cite{reisinger2007coevolving} not only evolved the network weights but also the architecture using NEAT algorithm~\cite{stanley2002evolving}. The evolved network was used as a state evaluator for the alpha-beta pruning algorithm~\cite{russell2010artificial} to play different board games from the General Game Playing Framework~\cite{genesereth2005general}. The best-evolved agent was able to beat the random agent in 5 different games from the General Game Playing framework. 
Finally, Gauci and Stanely~\cite{gauci2011evolving} used the HyperNEAT algorithm~\cite{stanley2009hypercube} to prune some branches for the Min-Max algorithm~\cite{russell2010artificial} besides the alpha-beta pruning. 
The final agent was able to have a higher win rate and more ties compared to the default alpha-beta pruning algorithm. 

We can notice that most of the work in that area is older than 10 years ago and we could not find any new work that combines EML with other search algorithms. We think that the boom in computation power and the dependence on the backpropagation algorithm is the main cause of that. For example, the Alpha-Go algorithm~\cite{silver2016mastering} uses a neural network as a state evaluator for MCTS agent~\cite{browne2012survey}. This is similar to the Blondie24 agent with the difference of having more computation power to train a big network using backpropagation~\cite{hecht1992theory}. This does not mean that we should abandon EML for the sake of backpropagation, but it should push us towards using EML in a smarter and different way that backpropagation cannot do. For example, we could try evolving a network that compresses the game state space such that a dynamic programming agent can play the game efficiently.


\section{Conclusion}

As demonstrated by the myriad examples in this chapter, game research has been enhanced greatly by evolution and ML. The combination of the two approaches provides particularly compelling tools for generating game agents and content; the two types of methods naturally complement each other's strengths and weaknesses. The future of this research area is exciting, as rapid advances in ML technologies will allow us to not only apply new algorithms to existing games but perhaps even to create new kinds of game-based challenges as well (which will, in turn, provide new domains for evaluating new kinds of EML systems).

\bibliographystyle{sn-basic}
\bibliography{sn-bibliography}


\end{document}